\documentclass[conference]{IEEEtran}
\IEEEoverridecommandlockouts
\usepackage{cite}
\usepackage{amsmath,amssymb,amsfonts}
\usepackage{graphicx}

\usepackage{algorithm}
\usepackage{algpseudocode}

\usepackage{textcomp}
\usepackage{xcolor}
\def\BibTeX{{\rm B\kern-.05em{\sc i\kern-.025em b}\kern-.08em
    T\kern-.1667em\lower.7ex\hbox{E}\kern-.125emX}}
\begin{document}

\title{An Adaptive Metaheuristic Framework for Changing Environments}

\author{\IEEEauthorblockN{Bestoun S. Ahmed}
\IEEEauthorblockA{\textit{Department of Mathematics and Computer Science,}
\textit{Karlstad University}
Karlstad, 65188, Sweden \\
Department of Computer Science, Faculty of Electrical Engineering, Czech Technical University in Prague\\ Prague, 16627, Czech Republic\\
bestoun@kau.se}
}

\maketitle

\begin{abstract}
The rapidly changing landscapes of modern optimization problems require algorithms that can be adapted in real-time. This paper introduces an Adaptive Metaheuristic Framework (AMF) designed for dynamic environments. It is capable of intelligently adapting to changes in the problem parameters. The AMF combines a dynamic representation of problems, a real-time sensing system, and adaptive techniques to navigate continuously changing optimization environments. Through a simulated dynamic optimization problem, the AMF's capability is demonstrated to detect environmental changes and proactively adjust its search strategy. This framework utilizes a differential evolution algorithm that is improved with an adaptation module that adjusts solutions in response to detected changes. The capability of the AMF to adjust is tested through a series of iterations, demonstrating its resilience and robustness in sustaining solution quality despite the problem's development. The effectiveness of AMF is demonstrated through a series of simulations on a dynamic optimization problem. Robustness and agility characterize the algorithm's performance, as evidenced by the presented fitness evolution and solution path visualizations. The findings show that AMF is a practical solution to dynamic optimization and a major step forward in the creation of algorithms that can handle the unpredictability of real-world problems.
\end{abstract}



\section{Introduction}
\label{sec:introduction}
The field of computational optimization is constantly evolving and dynamic. With the availability of many new applications, there is a need for algorithms that can adapt to changing environments. Traditional optimization techniques that have been designed mainly for static situations are not suitable for dynamic settings where parameters, constraints, and objectives are subject to continuous changes \cite{amaran2016simulation}. This has led to a greater focus on developing adaptive metaheuristic algorithms that respond to evolving problem landscapes in real-time. Metaheuristic algorithms are known for their flexibility and robustness, which makes them suitable for many complex optimization problems. However, their application in dynamic environments requires an additional layer of adaptability. The main challenge with adaptive metaheuristics is their ability to detect environmental changes and adjust their search strategies accordingly. The ultimate goal is to enable these algorithms to find optimal solutions and maintain or regain optimality as the problem evolves.


Recent advances in this field have been significant. Boks and Wang \cite{Boks2021} established a framework for the dynamic configuration of operators and parameters in Differential Evolution (DE). The research focused on combined fitness- and diversity-driven adaptation methods. Vakhnin, Sopov, and Semenkin \cite{math10224297} proposed an approach that combines multiple ideas from state-of-the-art algorithms, implementing Coordination of Self-adaptive Cooperative Co-evolution algorithms with Local Search (COSACC-LS1) for large-scale optimization problems. Martins et al. \cite{Martins2021} conducted a comparative analysis of metaheuristics applied to adaptive curriculum sequencing, demonstrating the effectiveness of Differential Evolution (DE) in this context. Penas et al. \cite{Penas2017} presented a parallel metaheuristic for large mixed-integer dynamic optimization problems with applications in computational biology that show the scalability and efficiency of these methods. It is clear from the literature that the development and application of these sophisticated algorithms have been largely designed to address specific types of optimization challenges. This specificity highlights the need for adaptive algorithms that not only are designed to work within particular application domains, but also show a degree of generalization that current solutions may lack. However, the purpose of such adaptiveness shows a critical gap in optimization problem solving, which is the absence of a universal framework capable of dynamically adjusting its components, be they operators, parameters, or problem-solving strategies, to suit a wide array of problem types without sacrificing efficiency or effectiveness.

Building upon these insights, this paper introduces the Adaptive Metaheuristic Framework (AMF), specifically designed for dynamic optimization problems. The AMF framework is designed to adjust its search process to any alterations in the problem environment smartly, with a variable problem representation that reflects the ever-changing nature of real-world issues. This is complemented by a real-time sensing mechanism that detects shifts in the problem environment and then prompts an adaptive response from the algorithm. The innovation of the AMF lies in its integration of a differential evolution algorithm with an adaptation module, which fine-tunes solutions in response to detected environmental changes. This ensures the relevance and efficacy of solutions in dynamic contexts. To avoid the specificity of this framework for a special domain, the performance of the framework is evaluated through simulations of a dynamic optimization problem. The evaluation shows its ability to maintain high-quality solutions despite frequent and unpredictable changes. Moreover, the evaluation also shows the algorithm's performance and detailed insights into the fitness evolution and the solution's trajectory through the search space.

The paper makes several significant contributions to dynamic optimization. The AMF is a robust and flexible strategy designed to adapt to changing environments, a critical feature for real-world optimization scenarios. It has a unique sensing mechanism and an adaptation strategy that allows it to detect environmental changes and adjust its search process accordingly. In addition, it incorporates advanced techniques, such as differential evolution and local search strategies, to react effectively to the evolving nature of optimization problems. The superior performance, scalability, and efficiency of the framework are demonstrated here through extensive experimental evaluations compared to traditional metaheuristic approaches. Another contribution of this work is its advances in the theoretical understanding of adaptive optimization strategies and its practical insights and tools for tackling complex and dynamic problems across various domains.

The rest of this paper is organized as follows to provide a comprehensive understanding of the AMF framework and its applications in dynamic optimization. Following the introduction, Section \ref{RelatedWork} discusses Related Work, offering a critical review of the existing literature and highlighting advances and gaps in the field. Section \ref{AdaptiveMechanism} gives an overview of the framework and elaborates on its core principles, structure, and operational dynamics. Section \ref{AlgorithmAdaptation} discusses the specific mechanisms that allow the AMF to adapt effectively to changing environments. Section \ref{Evaluation} is dedicated to a detailed empirical assessment of the AMF. A series of experiments are designed to test the framework and the results are discussed in detail. The paper concludes with Section \ref{Conclusion}, in which it summarizes the findings and reflects on the implications of this research and suggests future work directions in this exciting and evolving domain.


\section{Related Work}\label{RelatedWork}

Metaheuristic optimization has been steadily evolving, especially in dynamic contexts. Much effort has been put into adapting these algorithms to the ever-changing real-world scenarios. This development is evidenced by the numerous innovative techniques and applications presented in recent studies.

Valdez, Castillo, and Melin \cite{Valdez2021} investigated using bioinspired algorithms to optimize fuzzy clustering. Their findings demonstrate the effectiveness of nature-inspired optimization techniques in addressing complex problems, a concept reflected in many areas where traditional algorithms are inadequate. Similarly, Oladipo, Sun, and Wang \cite{Oladipo2020} studied the optimization of PID controllers with metaheuristic algorithms for DC motor drives, illustrating the practical applications of these algorithms in industrial settings, such as robotics and automotive engineering. Zitouni et al. \cite{Zitouni2022} further demonstrate a practical approach by introducing the Archerfish Hunting Optimizer, a novel metaheuristic algorithm that exhibits robustness and successful convergence in global optimization tasks. In this paper, multiobjective optimization is also used to deal with PID tuning problems in which multiobjectives are used to deal with the change in parameter tuning \cite{sahib2016new}. However, increasing the objectives with significant dynamic problems will not lead to a scalable solution. In addition, there is a need to adjust the objective manually with the availability of new conditions.

Osaba et al. \cite{Osaba2021} have created a thorough tutorial on the design and utilization of metaheuristics for real-world optimization issues. The tutorial emphasizes the necessity of accuracy and openness in algorithm development, which is essential for the advancement of the field. Uzor et al. \cite{Uzor2014} also highlighted the significance of real-world assessments over artificial benchmarks by introducing an adaptive-mutation compact genetic algorithm evaluated using dynamic optimization problems.

Rajwar, Deep, and Das \cite{Rajwar2023} comprehensively examined metaheuristic algorithms, tracking around 540 of them, providing a comprehensive taxonomy, and discussing their uses and difficulties. This extensive review highlights the rapid growth and broad use of metaheuristics in various areas, emphasizing the need for new and effective optimization techniques. Similarly, Game, Vaze, and Emmanuel \cite{Game2020} discussed the importance of bioinspired optimization algorithms to tackle complex real-world problems, particularly large-scale industrial and engineering issues. Their insights into the adaptability and efficiency of these algorithms in avoiding local optima and finding global solutions are essential to understanding the potential of metaheuristics in practical applications. 

Kumar, Das and Zelinka \cite{Kumar2020} made a notable contribution to the field of optimization with their new variant of the Spherical Search (SS) algorithm. This version of the algorithm features a self-adaptation structure, which greatly improves its performance. The algorithm was tested on 57 real-world optimization problems, showing its effectiveness and efficiency in tackling complex, non-convex optimization tasks. The success of this self-adaptive SS algorithm is a testament to the increasing sophistication and specialization in the design of metaheuristic algorithms. It reflects the trend towards creating more intelligent and responsive optimization tools, which can handle the complexities of real-world applications.

Yi \cite{Yi2017} has demonstrated the potential of hybrid metaheuristic algorithms in intelligent architectural design decisions by integrating them into adaptive real-time optimization. This approach to parametric shading design, which combines wireless data transfer equipment and a parametric visual programming language, is an example of the innovative uses of these algorithms beyond traditional optimization problems.

This paper contributes to the field of metaheuristic optimization algorithms by introducing the AMF Framework. Unlike other studies in the literature that introduce algorithms for special application domains, this paper introduces a framework that is designed to be adaptive, allowing it to sense and respond to real-time changes, giving a better understanding of the algorithm's behavior in dynamic environments. The framework uses a dynamic simulation of ever-changing applications, which is not tied to specific applications and guarantees its generalizability for many applications in a changing environment. This paper adds to the growing consensus that metaheuristic optimization algorithms must be adaptive and capable of handling the complexities of real-world problems.


\section{Overview of the Adaptive Metaheuristic Framework (AMF)}\label{OverviewAMF}

The AMF framework provides a mathematical formulation for dynamic optimization problems (DOPs) that captures the changing nature of real-world scenarios. A DOP is characterized by an objective function $f(x,t)$, where $x$ is the vector of decision variables and $t$ is the time or state of the environment, reflecting the dynamic nature of the problem. This objective function is not static, but instead changes over time, reflecting the ever-changing nature of real-world challenges.

In addition to the objective function, the DOP includes a set of constraints $g(x, t) \leq 0$ and $h(x,t) =0$ where $g$ and $h$ represent inequality and equality constraints, respectively. These constraints are also dependent on the state of the environment, making the problem even more complex. The decision variables $x$ are usually confined within a feasible search space, which is determined by lower and upper bounds $x_(min)$ and $x_(max)$. This bounded search space guarantees that the solutions remain practical and within the boundaries of real-world applicability.

The goal of the AMF framework is to identify the optimal solution $x^*$ through its optimization algorithm. The optimal solution maximizes or minimizes the objective function $f(x,t)$ while satisfying the constraints $g(x,t)$ and $h(x,t)$ in all time periods or states. To achieve optimal results, the framework must be able to identify and produce high-quality solutions and modify and customize them over time as the problem at hand undergoes changes. This involves continuously monitoring and analyzing the effectiveness of existing solutions and then applying the necessary modifications and improvements to ensure continued success. Only by adapting to evolving challenges and circumstances can the framework deliver consistent and sustainable outcomes. The dynamic nature of the objective function and the constraints implies that the optimal solution $x^*$ at one time point may not remain optimal as the environment evolves. Therefore, the optimization algorithm within the AMF continually modifies its search strategies in response to these changes, guaranteeing that the solutions remain pertinent and effective.

The AMF framework is a significant achievement in computational optimization, particularly when resolving issues caused by dynamic environments or testing which optimization algorithm performs better in changing environments. It has been specifically designed to adapt to real-time changes in the optimization landscape, guaranteeing the relevance and efficiency of its results. The AMF is made up of several important elements, all of which are crucial for its adaptability. These fundamental components are as follows.


\begin{enumerate}
    \item \textbf{Dynamic Problem Representation:} AMF utilizes a dynamic problem representation model at its core, which can replicate the constantly changing nature of real-world issues. This model accurately reflects the varying constraints, objectives, and parameters of the problem. By capturing the core of dynamic conditions, AMF can be used to show how quickly the used algorithm can adapt to modifications and ensure that the optimization process remains in sync with the current state of the problem.

    \item \textbf{Real-Time Sensing Mechanism:} The AMF has a real-time sensing system constantly monitors the environment for any changes. This system can detect any alterations or variations in the parameters and restrictions. Whenever a shift is identified, the sensing system triggers the adaptive response of the framework, which ensures that the optimization approach remains optimal in the new circumstances.

    \item \textbf{Adaptive Optimization Algorithms:} At the heart of the AMF lies a suite of adaptive optimization algorithms. These algorithms are designed to be highly flexible, allowing them to adjust their search strategies based on the data collected by the sensing system. This adaptability is crucial to ensure optimal solutions, even when the problem environment undergoes frequent and unexpected changes. The current paper uses the Differential Evolution (DE) algorithm for optimization. However, the algorithm can be replaced with other suitable algorithms for testing in the framework. 


    \item \textbf{Adaptation Module:} The framework is a unique system that combines the optimization algorithm (DE algorithm here) with an adaptation module. This module adjusts the solutions to any environmental changes, ensuring that the solutions remain effective and applicable. The DE algorithm is renowned here for its dependability and effectiveness, and with the addition of adaptive capability, it is especially suitable for dynamic optimization tasks.

    \item \textbf{Performance Evaluation:} Simulations of dynamic optimization problems are used to evaluate the effectiveness and robustness of the AMF. These simulations are crucial to demonstrate that the framework can produce high-quality solutions despite the dynamic nature of the problems. Based on the performance evaluation results of these simulations, the framework is improved to ensure its performance and dependability.

\end{enumerate}

\begin{figure}
    \centering
    \includegraphics[width=0.7\linewidth]{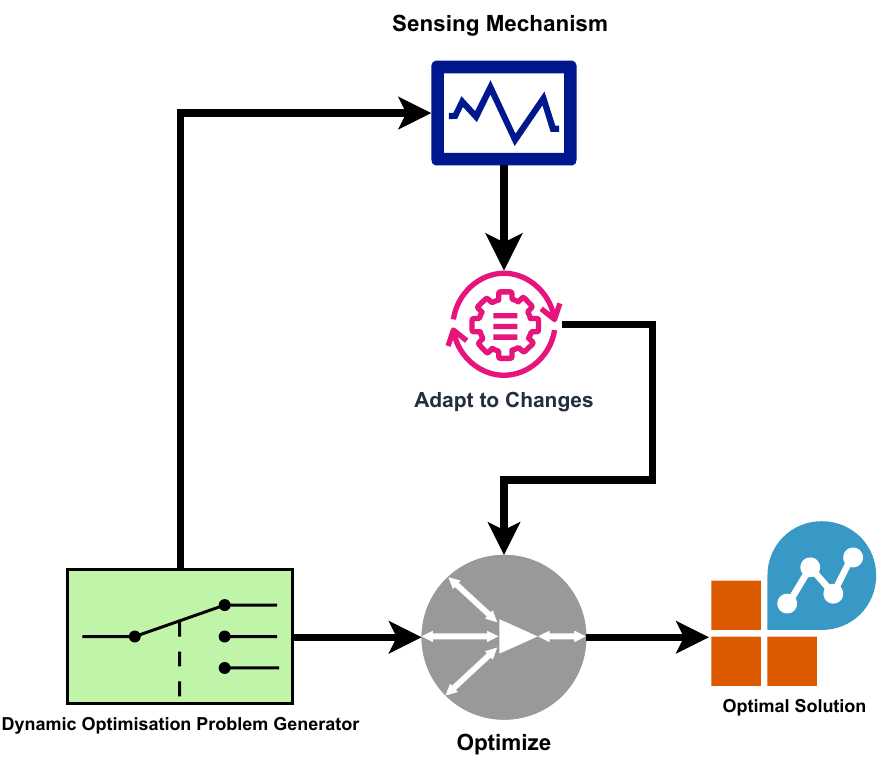}
    \caption{AMF framework workflow}
    \label{fig:AMFWorkflow}
\end{figure}

The AMF framework workflow is illustrated in Figure \ref{fig:AMFWorkflow}. The problem generator creates optimization problems with various parameters, constraints, and objectives. This ensures that the optimizer produces optimized solutions that cater to varying requirements. The optimizer utilizes the generated problem to produce an optimized solution. The sensing mechanism module is responsible for the ongoing monitoring of any modifications in the problem generator, allowing for immediate detection of any changes. Upon detection of a change, the adaptation trigger is activated, signaling the optimizer to adapt to the revised problem. The adaptation trigger ensures that the optimizer produces an optimized solution that caters to the latest problem requirements. Overall, the AMF framework workflow ensures that the optimizer produces optimized solutions that cater to varying requirements and detect any changes in the problem generator in real-time.

\section{Differential Evolution (DE) Algorithm}

This paper uses DE as a reliable algorithm within the framework. DE is a powerful stochastic optimization algorithm widely used to solve complex optimization problems. It works by maintaining a population of candidate solutions and iteratively improving them using a set of predefined rules. DE was first introduced by Storn and Price in 1997 \cite{storn1997differential} and has since become popular due to its simplicity, efficiency, and robustness. It is particularly effective in solving global optimization problems with many variables, nonlinearity, and noise. 

DE is used within this framework because it has several advantages over other optimization algorithms. First, it does not require the objective function to be differentiable, which makes it suitable for a wide range of problems, such as dynamic problems. Second, it is computationally efficient and can handle problems with a large number of variables. Third, it is robust to noise and outliers, making it suitable for real-world applications where data may not be perfect.


DE is based on the principle of mutation and crossover. Mutation is a random perturbation of the candidate solutions that introduces diversity into the population. Crossover is a process of combining two or more candidate solutions to create new ones. The combination is done in such a way that the resulting solution inherits the best characteristics of its parents. To this end, the DE algorithm can be formalized as follows: let $\mathbf{P}=\left\{\mathbf{x}_1, \mathbf{x}_2, \ldots, \mathbf{x}_N\right\}$ be the population of candidate solutions. For each generation $g$, perform the following steps for each $\mathbf{x}_i \in \mathbf{P}$:

\begin{itemize}
    \item \textbf{Mutation:} 

    $\mathbf{v}_i^g=\mathbf{x}_{r 1}^g+F \cdot\left(\mathbf{x}_{r 2}^g-\mathbf{x}_{r 3}^g\right)$

    \item \textbf{Crossover:}

    $\begin{cases}v_{i j}^g & \text { if } \operatorname{rand}(j) \leq C R \text { or } j=\operatorname{rand}(D) \\ x_{i j}^g & \text { otherwise }\end{cases}$

    \item Selection:

    $\begin{cases}\mathbf{u}_i^g & \text { if } f\left(\mathbf{u}_i^g\right) \leq f\left(\mathbf{x}_i^g\right) \\ \mathbf{x}_i^g & \text { otherwise }\end{cases}$

\end{itemize}


During the initialization phase of the algorithm, a population of $N$ candidate solutions (vectors) is generated randomly. Each vector $\mathbf{x}_i=\left(x_{i 1}, x_{i 2}, \ldots, x_{i D}\right) \text { for } i=1,2, \ldots, N$ is initialized, where $D$ is the dimensionality of the problem. During the mutation phase, for each vector $x_i$, a mutant vector $v_i$ is generated. The mutation is performed using the formula $\mathbf{v}_i=\mathbf{x}_{r 1}+F \cdot\left(\mathbf{x}_{r 2}-\mathbf{x}_{r 3}\right)$. Here, $r1$, $r2$, and $r3$ are distinct indices randomly chosen from the population, and $F$ is a scaling factor. On the other hand, the crossover operation creates a trial vector $\mathbf{u}_i$ by mixing the mutant vector $v_i$ with the target vector $x_i$. A crossover rate $CR$ determines the probability that elements of the mutant vector are passed to the trial vector. The trial vector $\mathbf{u}_i$ is evaluated during the selection phase, and its fitness is compared with the target vector $x_i$. If $\mathbf{u}_i$ has equal or better fitness, it replaces $x_i$ in the population; otherwise, $x_i$ is retained.

The adaptive framework uses DE as a local search method to adapt to the changes in the optimization landscape. Whenever a change is detected in the environment, DE is applied to quickly find a new, near-optimal solution in the area of the current solution. This approach benefits from DE's ability to efficiently explore and exploit the solution space, making it an ideal choice for dynamic optimization problems. Integrating DE into the adaptive framework enhances its ability to respond to environmental changes, ensuring a consistently high level of solution quality throughout the optimization process. This synergy between the adaptive framework and DE's search capabilities forms the foundation of the current approach to addressing dynamic optimization challenges.


\section{Algorithm Adaptation Mechanism }\label{AlgorithmAdaptation}

The Algorithm Adaptation Mechanism is one of the most advanced features of the AMF framework. It ensures that the framework can quickly adapt to any changes in the optimization landscape, which is critical for efficient and relevant optimization processes. The mechanism operates through crucial steps that enable it to make informed decisions based on real-time data. First, it identifies any changes in the optimization landscape, such as the emergence of new challenges or shifts in the problem environment. Then, it evaluates the impact of these changes on the optimization process and determines whether any adjustments are necessary. If adjustments are required, the mechanism automatically generates new algorithms or modifies existing ones to address the changes in the landscape. Finally, it tests these new algorithms to ensure that they are effective and efficient before incorporating them into the optimization process. The following bullets capture the essential parts of the algorithm:

\begin{itemize}
    \item \textbf{Monitoring and Detection:} This mechanism constantly monitors the issue environment, using real-time sensors to detect changes in the objective function, constraints, or other relevant parameters.
    
    \item \textbf{Assessment and Decision Making:} When a change is identified, the system evaluates the type and magnitude of the change. On the basis of this evaluation, it makes informed choices about modifying the optimization approach.
    
    \item \textbf{Adaptation of the Search Strategy:} The adaptation mechanism is based on its ability to adjust the search strategy of the metaheuristic algorithm. This could include changing parameters, balancing exploration and exploitation, or substituting the algorithmic approach if needed.
    
    \item \textbf{Feedback and Learning:} This mechanism includes a feedback loop in which the results of the adjusted strategy are examined to guide future modifications. This learning element is essential to improve the adaptation process in terms of efficiency and effectiveness over time.
\end{itemize}

Algorithm 1 shows the necessary steps of the framework.

\begin{algorithm}\label{AMFalgorithm}
\caption{Adaptive Metaheuristic Framework (AMF)}
\begin{algorithmic}[1]
\State Initialize AMF with a population of solutions
\State Define objective function $f(x, t)$
\State Define constraints $g(x, t) \leq 0$ and $h(x, t) = 0$
\State Set bounds for decision variables $x_{\text{min}}$ and $x_{\text{max}}$

\While{termination criteria not met}
    \State $t \gets \text{getCurrentTime()}$ or $\text{getState()}$
    
    \For{each solution $x$ in population}
        \State Evaluate $f(x, t)$ subject to $g(x, t)$ and $h(x, t)$
    \EndFor
    
    \If{changeDetected($t$)}
        \For{each solution $x$ in population}
            \State Adapt solution $x$ based on detected change
            \State Adjust algorithm parameters if necessary
        \EndFor
    \EndIf
    
    \State Apply metaheuristic search strategy to population
    \State Update population based on search results
    
    \If{feedbackAvailable()}
        \State Learn from feedback and adjust adaptation strategy
    \EndIf

\EndWhile

\State \Return best solution $x^*$ found
\end{algorithmic}
\end{algorithm}

In the beginning, the framework establishes a population of potential solutions. It defines the objective function $f(x,t)$, the constraints $g(x, t) \leq 0$ and $h(x,t)=0$, and the boundaries for the decision variables $x_{\text{min}}$ and $x_{\text{max}}$. The algorithm's main loop continues until a predetermined termination criterion is met. During this loop, the current time or state $t$ is determined, which is essential to address the dynamic aspect of the problem. The objective function is evaluated under the current constraints for each solution in the population. If a change is detected in the problem environment, each solution is adapted accordingly, and the parameters of the algorithm may be adjusted to better suit the new conditions. The metaheuristic search strategy is then applied to the population, followed by an update of the population based on the results of this search. Furthermore, suppose that the available feedback is used to refine and improve the adaptation strategy. In that case, this process is repeated iteratively, constantly adapting to changes until the best solution $x^*$ is found and returned. This pseudocode encapsulates the adaptive and responsive nature of the AMF, demonstrating its capacity to dynamically adjust to changing environments in the pursuit of optimal solutions.

The adaptive optimization Algorithm is designed to tackle dynamic optimization problems by intelligently adapting to changing environments, as shown in Algorithm 2.

\begin{algorithm*}\label{AdaptiveMechanism}
\caption{Adaptive Optimization for Dynamic Optimization}
\begin{algorithmic}[1]
\State \textbf{Initialize:}
\State Define a dynamic optimization problem with dimensions, change frequency, and change severity.
\State Create a sensing mechanism to detect changes in the environment.
\State Set up adaptation strategies for responding to environmental changes.
\State Initialize other components such as memory incorporation, learning component, real-time evaluation, self-adjustment, and multi-agent approach.

\State \textbf{Define the Dynamic Optimization Problem:}
\State Initialize the optimal solution $x^*$ at random.
\State Define an evaluation function $f(x, t)$ that changes the optimal solution at specified intervals (change frequency).

\State \textbf{Sensing Mechanism:}
\State Implement a function to evaluate the current solution $x$ and detect environmental changes.

\State \textbf{Adaptation Strategy:}
\State Define a strategy to adapt to the sensed changes, such as re-initializing part of the population or performing a local search around the current solution.

\State \textbf{Main Loop (for a specified number of iterations):}
\While{termination criteria not met}
    \State Evaluate the current solution $x$ using the sensing mechanism.
    \If{change detected (based on change frequency)}
        \State Adapt the current solution $x$ using the adaptation strategy.
    \EndIf
\EndWhile

\State \textbf{Visualization and Analysis:}
\State Collect data for visualization (e.g., fitness over time, solution path).
\State After optimization, visualize the fitness evolution, solution path, and other relevant metrics.

\State \textbf{Feedback and Fine-tuning:}
\State Implement a feedback loop to fine-tune the algorithm's behavior based on performance.

\end{algorithmic}
\end{algorithm*}

The algorithm defines the problem's parameters, such as its size and the frequency and intensity of environmental changes, and sets up a sensing system to detect them. It also prepares various adaptation strategies to respond to the sensed changes and initiates additional components such as memory, learning, real-time evaluation, self-adjustment, and a multi-agent system. The core of the algorithm involves formulating a dynamic optimization problem with a randomly initialized solution and an evaluation function that mimics the changing nature of the problem. The algorithm's main loop continually evaluates the current solution, adapting it when changes are detected, and optionally incorporates learning and self-adjustment mechanisms to improve its performance. Throughout the process, data are collected for post-optimization analysis and visualization, providing insights into the algorithm's effectiveness. Finally, a feedback loop is established to continuously refine and tune the algorithm based on its performance, ensuring its adaptability and robustness in dynamic and complex optimization scenarios.


\section{Experimental Evaluation and Discussion}\label{Evaluation}

This section attempts to conduct a comprehensive set of tests to evaluate the performance, flexibility, and scalability of the AMF framework in dynamic optimization scenarios. These assessments were crucial in demonstrating the effectiveness of the AMF and its potential to be utilized in various real-world situations. The initial experiments examined the effectiveness of AMF in a simulated dynamic optimization task. It was observed that the framework could adjust to alterations in the problem landscape, maintaining high-quality solutions despite frequent and unpredictable changes in the problem parameters. This adaptability was further demonstrated through visualizations that showed the evolution of fitness and the path of the solution through the search space. These visualizations provided proof of the AMF's ability to manage the complexities of dynamic environments effectively.

The AMF was compared with other population-based optimization algorithms to assess its performance. It was found to be superior in terms of the quality of the solutions it produced and its ability to adapt to changing conditions. This comparison highlighted the AMF's superior design for dealing with dynamic optimization problems, distinguishing it from traditional methods.

\begin{figure}
    \centering
    \includegraphics[width=1\linewidth]{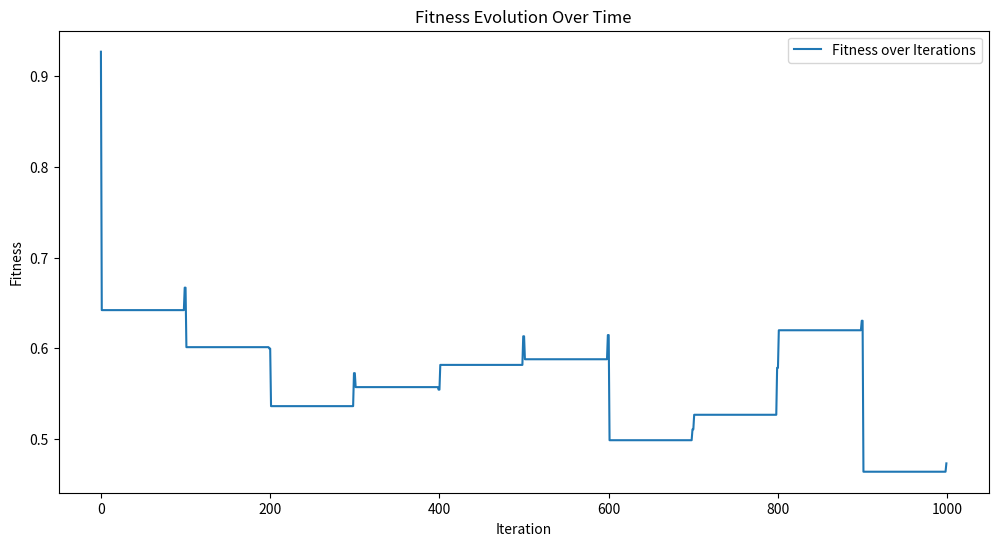}
    \caption{Comparison of Fitness function evaluation over time with iteration increase}
    \label{fig:fitnessEvaluation}
\end{figure}

The evaluation begins by showing how the fitness function is adjusted over time and compares the fitness function evaluation with the increase in iterations. Figure \ref{fig:fitnessEvaluation} shows the result of this evaluation.

\begin{figure}
    \centering
    \includegraphics[width=1\linewidth]{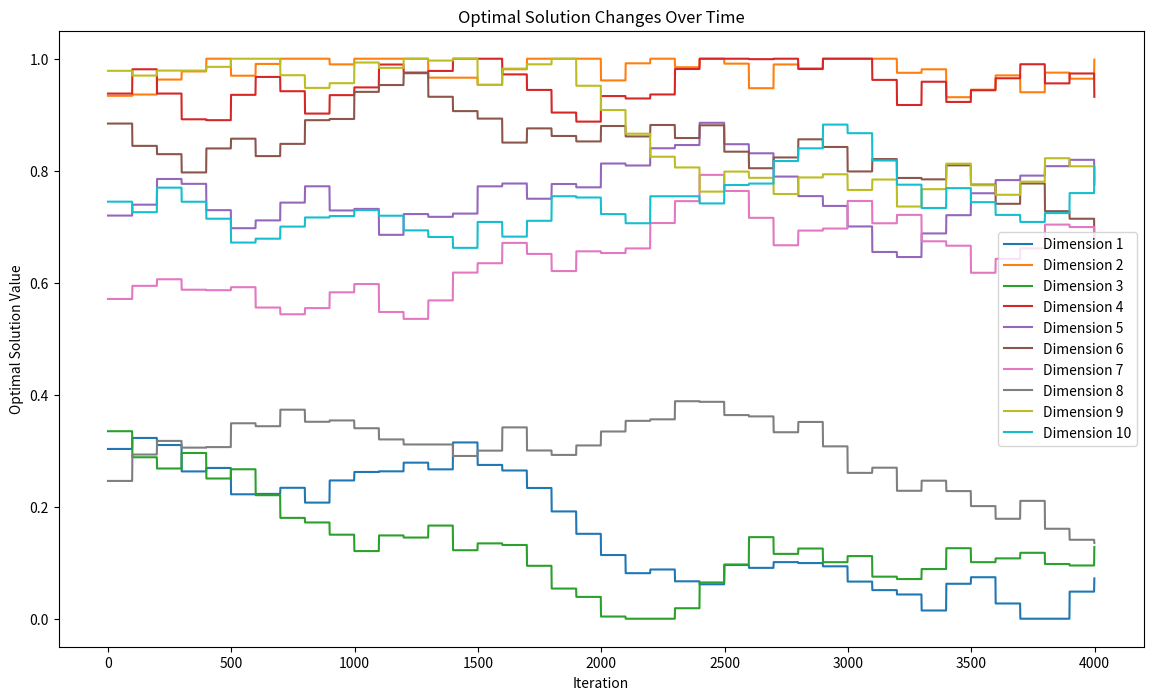}
    \caption{Evaluation of how the fitness function of the problem change over time}
    \label{fig:FitnesWithDimentions}
\end{figure}

The results in Figure \ref{fig:fitnessEvaluation} is a line graph that tracks the fitness of the best solution found in each iteration of the optimization process. The line plot represents the change in fitness over time. Each point on the line corresponds to the fitness of the best solution in a particular iteration. The figure tracks the change in a fitness value over a series of iterations, from 0 to 1000. In the graph, it is clear that there is a sharp decrease in fitness from the first iteration, suggesting that the optimization method found a significantly better solution early on. The dynamic problem generator changes the nature of the problem by altering the parameters and constraints, as well as the objective function, in each 200 iterations. It is clear from the graph that the sensing mechanism can detect this change, and the optimization algorithm adapts to the change and changes the fitness function accordingly. To better show how the framework operates with different problem dimensions, 10 problem dimensions have been defined and how the algorithm effectively adapts and deals with each dimension. Figure \ref{fig:FitnesWithDimentions} shows the result of these different dimensions of the problem and the performance of the algorithm. The x-axis, labeled "Iteration," shows the progression of the optimization algorithm over time, and the y-axis, labeled "Optimal Solution Value," represents the quality of the solution found.

Figure \ref{fig:FitnesWithDimentions} presents the evolution of the optimal solution values in ten different dimensions over a series of iterations (from 0 to 4000). Each line represents one of the ten dimensions, each color corresponding to a different dimension. The lines show how the value of the optimal solution for each dimension changes as the optimization process progresses. The graph shows that different dimensions stabilize at different levels of optimal values, indicating that the search space is complex and that each dimension is being optimized to a different degree.

\begin{figure}
    \centering
    \includegraphics[width=1\linewidth]{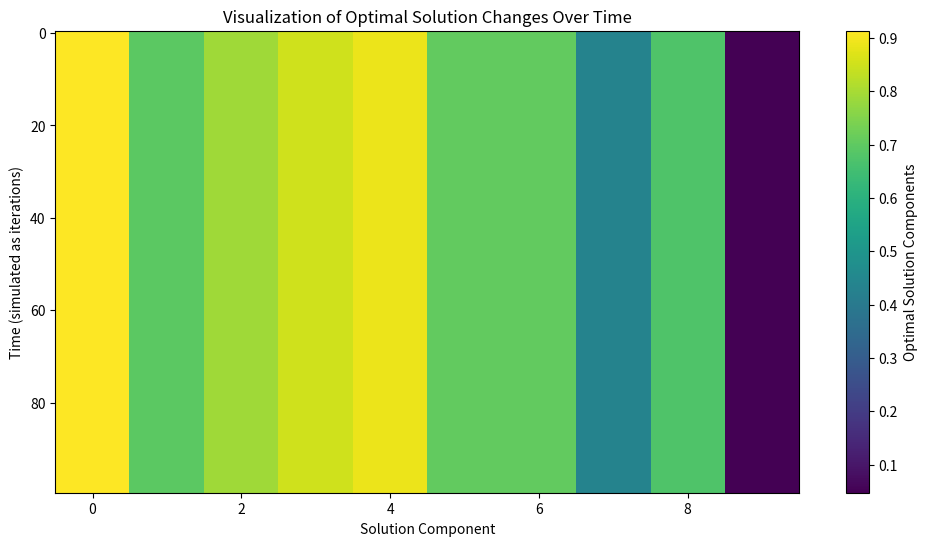}
    \caption{Optimal Solution Changes over time}
    \label{fig:FitnessHeatmaps}
\end{figure}

To evaluate how the dynamic optimization problem generator generates problems effectively, Fitgure \ref{fig:FitnessHeatmaps} displays the change in the components of the optimal solution over several iterations. The x-axis represents different solution components in a multidimensional optimization problem, while the y-axis represents discrete time steps simulated as iterations in the optimization process.

In Figure \ref{fig:FitnessHeatmaps}, the intensity of the color represents the value of the optimal solution components, with the scale shown on the right side of the heat map ranging from 0 to 0.9. The colors change from yellow (higher values) to dark purple (lower values), suggesting a gradient of values for the optimal solution at each component over time. For instance, earlier iterations (toward the top of the heatmap) show higher values (yellow), and as time progresses (moving downward), the values for each component either remain constant or change.

\begin{figure}
    \centering
    \includegraphics[width=1\linewidth]{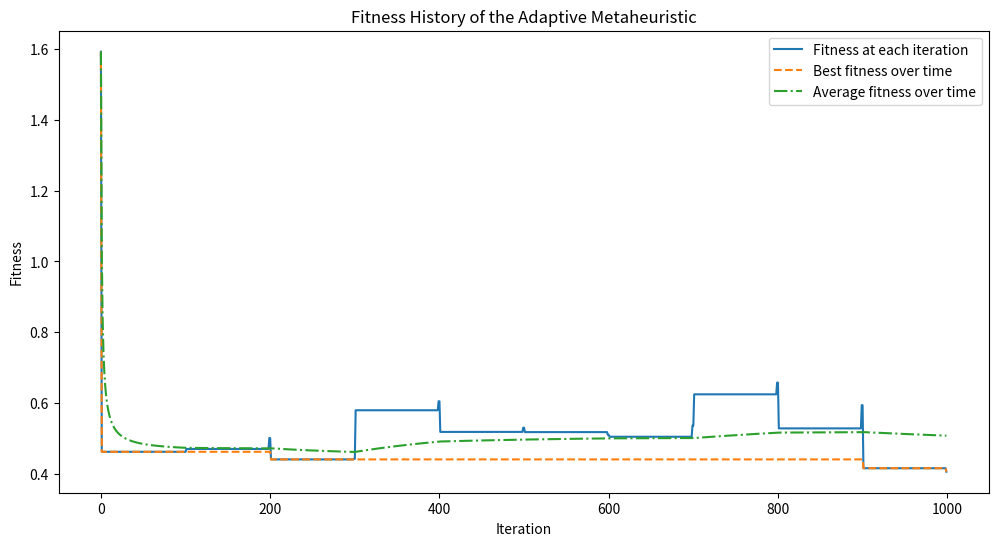}
    \caption{Fitness History of the Adaptive Algorithm}
    \label{fig:FitnessHistory}
\end{figure}

To fully understand the workings of the adaptation algorithm with the fitness function and how it adapts to the problem generator, three different aspects of fitness are comprehensively compared. These aspects include fitness in each iteration, the best fitness over time, and the average fitness over time. The analysis was carried out from iteration 0 to 1000. This analysis was designed to gain a deeper understanding of how the algorithm interacts with the fitness function at different stages of the optimization process. The results of the analysis are presented in Figure \ref{fig:FitnessHistory}.

The results in Figure \ref{fig:FitnessHistory} show the fitness at each iteration in the solid blue line. This line shows the fitness of the potentially best solution found in each iteration. A steep drop in initial iterations indicates that the algorithm quickly found a much better solution early on. After this drop, the fitness value fluctuates slightly but remains low overall, suggesting that the algorithm is refining the solution. The best fitness over time is also shown in the figure with a dashed orange line. This line represents the best solution found at each point in time. It flattens out after a rapid improvement initially, indicating that the algorithm did not find a better solution than the initial one as the iterations progressed. The average fitness over time is shown in the dashed green line. This line tracks the average fitness of all solutions considered over time. The average fitness drops quickly at the beginning, along with the best fitness, and then remains relatively constant, indicating that, on average, the population of solutions does not improve much after the initial iterations. Based on the graph, it can be observed that the adaptive metaheuristic algorithm utilized efficiently found a good solution quickly. However, it encountered challenges in obtaining substantial improvements after the initial stages. The proximity of the best and average fitness lines after the initial decline implies that the algorithm maintains a stable population of solutions with comparable fitness values.

\begin{figure*}
    \centering
    \includegraphics[width=0.7\linewidth]{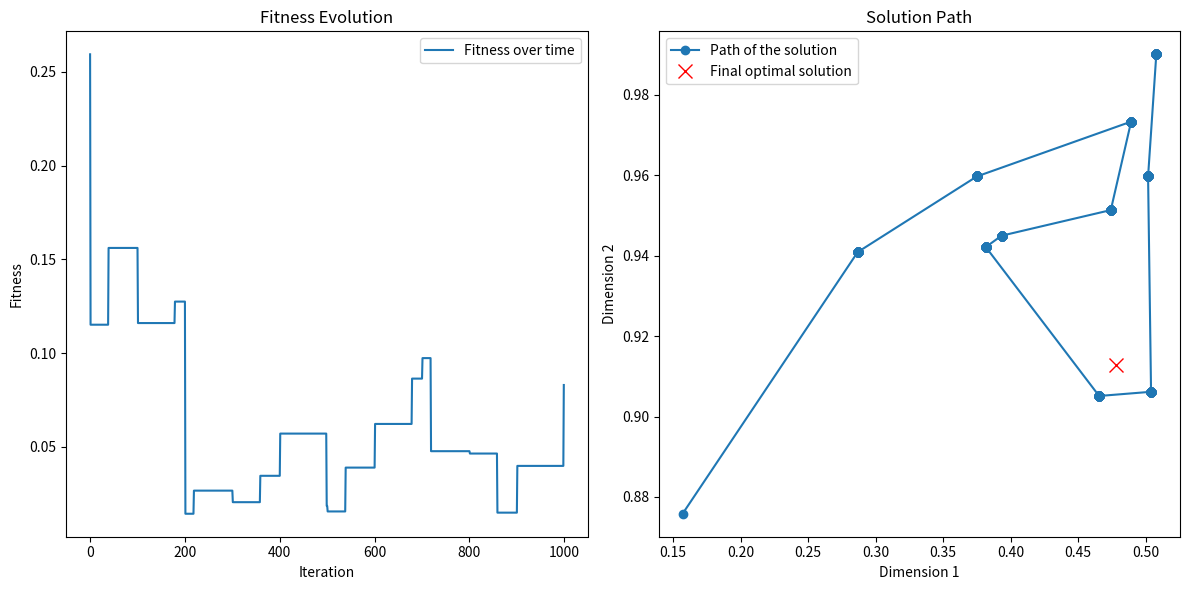}
    \caption{The trajectory path of the solution}
    \label{fig:SolutionPath}
\end{figure*}

The fitness function was monitored with the interaction to show how the algorithm reaches each solution over time with the dimension of the problem. Figure \ref{fig:SolutionPath} shows the results of this evaluation.

The results in Figure \ref{fig:SolutionPath} show the trajectory of the current solution through the search space throughout the optimization. The scatter plot with a connected line shows the path the solution has taken through the search space. Each point represents the solution in a specific iteration, with lines connecting them to show the path over time. The Red 'X' Marker Indicates the final optimal solution found at the end of the optimization process. The X-axis ('Dimension 1') and Y-axis ('Dimension 2') represent the values of the first two dimensions of the solution space. 

These graphs give a comprehensive view of the optimization process. The left graph provides a macro view of the overall improvement in solution quality over time, while the right graph offers a micro view of the specific trajectory through the solution space to arrive at the best solution found by the algorithm. This is an indicative that the algorithm balances exploration and exploitation to navigate towards the optimum in a multi-dimensional landscape.

\begin{figure*}
    \centering
    \includegraphics[width=0.99\linewidth]{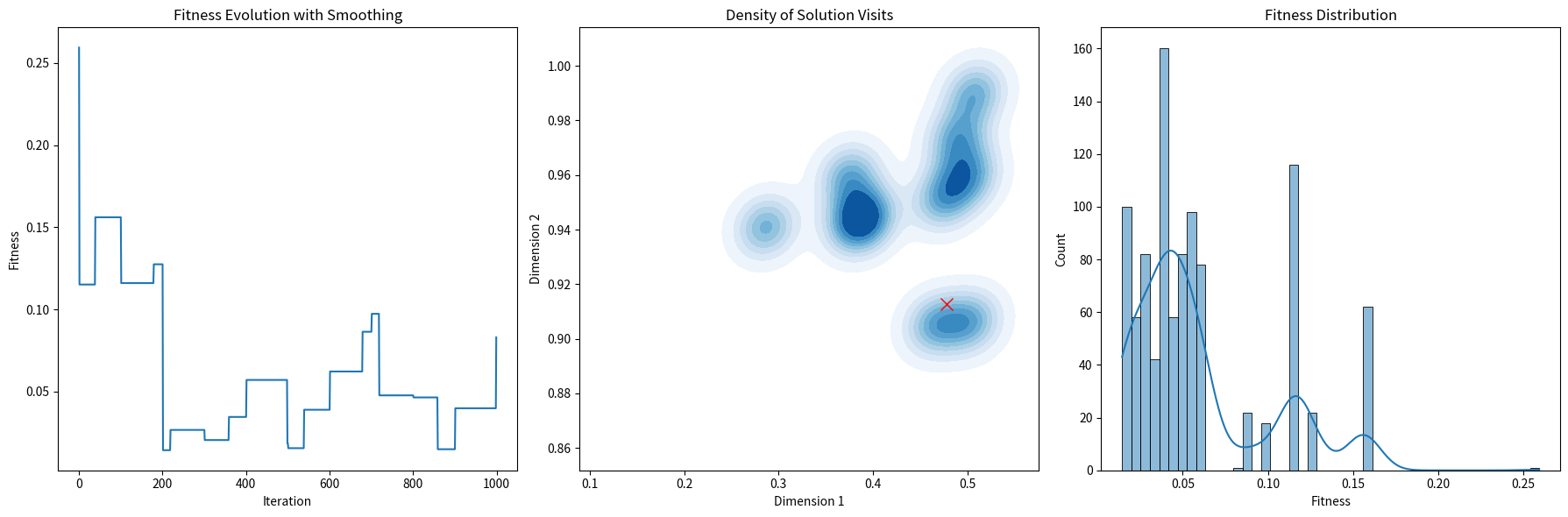}
    \caption{Comprehensive Analysis of Optimization Algorithm Performance}
    \label{fig:ComprehensiveAnalysisThreeGraph}
\end{figure*}

To further show the optimization process in one instance of a problem with various aspects, Figure \ref{fig:ComprehensiveAnalysisThreeGraph} shows the result of this analysis. These graphs provide a comprehensive overview of the behavior of the optimization algorithm. The first graph shows how solution quality evolved over time, the second graph shows where the algorithm searched and what it focused on, and the third graph indicates the variety and distribution of solutions' fitness values encountered throughout the optimization process. These results are useful in understanding the efficiency and characteristics of the optimization algorithm, including its exploratory behavior and convergence patterns.

The "Fitness Evolution with Smoothing" graph depicts fitness values over iterations, with a decreasing trend in fitness indicating improvement in the solution quality over time with the change of the problem environment behavior provided by the problem generator. The graph shows sharp declines in fitness at specific iterations due to the algorithm finding significantly better solutions at those points. The overall downward trend suggests that the optimization algorithm effectively reduces fitness, typically desired in optimization problems.

The "Density of Solution Visits" is a contour plot representing the frequency of the algorithm's visits to different solutions in a two-dimensional space defined by Dimensions 1 and 2. Darker areas indicate a higher visitation density, meaning that the algorithm has explored and utilized solutions in these regions more frequently. The red 'X' marks the final optimal solution found in one of the high-density areas, implying that the algorithm converged on this region as the search progressed.

The "Fitness Distribution" graph is a histogram overlaid with a smooth curve, representing a kernel density estimation showing the distribution of fitness values the algorithm has encountered. The x-axis represents the fitness values, and the y-axis shows how often these fitness values were observed. The distribution of fitness values has several peaks, indicating that the algorithm encountered multiple promising regions in the search space with similar fitness values.


To further analyze the performance of the DE algorithm with different settings, two different adaptation strategies were implemented and evaluated to assess their effectiveness in responding to dynamic changes in the optimization environment. These strategies are designed to modify the current solution in response to detected changes, aiming to maintain or improve the solution's fitness in the altered landscape.

\textbf{Strategy 1: Partial Re-initialization}

An effective strategy for optimizing solutions is "Partial Re-initialization". This approach is simple but effective. Whenever the environment changes, this strategy randomly re-initializes a portion of the current solution. Specifically, 10\% of the solution's components are reset with new random values. This method introduces fresh genetic material into the solution, which can help avoid local optima and explore new regions of the search space. The effectiveness of this strategy is due to its ability to find a balance between exploration and exploitation. Only a small part of the solution is altered, while the rest of the structure of the solution is preserved.

\textbf{Strategy 2: Local Search with Increased Mutation Rate}

The second strategy, named "Local Search with Increased Mutation Rate," adopts a more refined approach by utilizing the DE algorithm with a key modification: an increased mutation rate. The higher mutation rate is intended to intensify the search process, allowing for a more detailed exploration of the solution space surrounding the current solution. This strategy employs the 'best1bin' approach of DE \cite{narloch2017diversification} with a mutation factor range of 0.7 to 1.2, which is higher than the standard settings. This approach is particularly useful in dynamic environments where the landscape changes frequently, enabling the algorithm to adapt more aggressively to new conditions.

\begin{figure}
    \centering
    \includegraphics[width=1\linewidth]{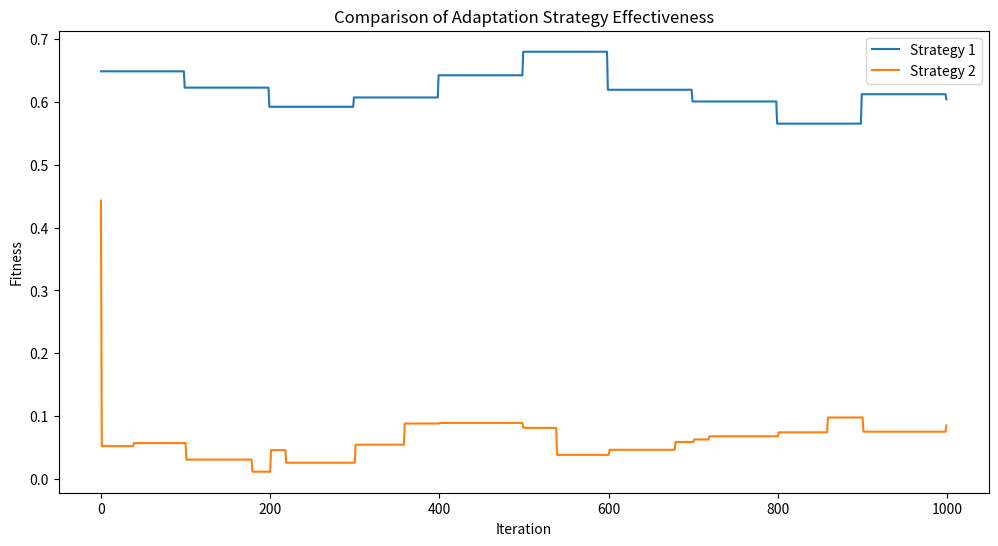}
    \caption{Comparative Performance of Adaptation Strategies in Dynamic Optimization}
    \label{fig:Strategy1and2Comparison}
\end{figure}

A series of experiments were conducted on the same dynamic optimization problem to evaluate the effectiveness of different strategies. The fitness of each strategy's solutions was measured over a set number of iterations to determine their performance. A visual comparison was made by plotting the fitness history of each strategy. The results provide insights into how each strategy copes with environmental changes. The performance of Strategy 1 reflects its ability to introduce variability while retaining the structure of the solution. In contrast, the performance of Strategy 2 indicates its ability to explore the solution space aggressively. This can be advantageous in rapidly changing environments. Figure \ref{fig:Strategy1and2Comparison} shows the detail of this evaluation for both strategies.

The results in Figure \ref{fig:Strategy1and2Comparison} show that Strategy 1 (re-initialization) has a pattern of sharp declines in fitness followed by gradual increases. The sharp declines correspond to the points where the strategy re-initializes part of the solution, which can quickly lead to improvements if the new random components are closer to the new optimal solution. However, the gradual increases in fitness suggest that this strategy struggles to maintain good solutions as the environment continues to change, possibly due to a lack of directed search after re-initialization. As for Strategy 2 (Local Search with High Mutation), the figure shows that it exhibits a more stable and consistent decline in fitness over time. The increased mutation rate allows the strategy to explore the search space more broadly, which is effective in finding and maintaining good solutions even as the environment changes. The smoother curve suggests that this strategy is more robust to environmental changes, likely due to its ability to explore new areas of the search space effectively.

\begin{figure}
    \centering
    \includegraphics[width=1\linewidth]{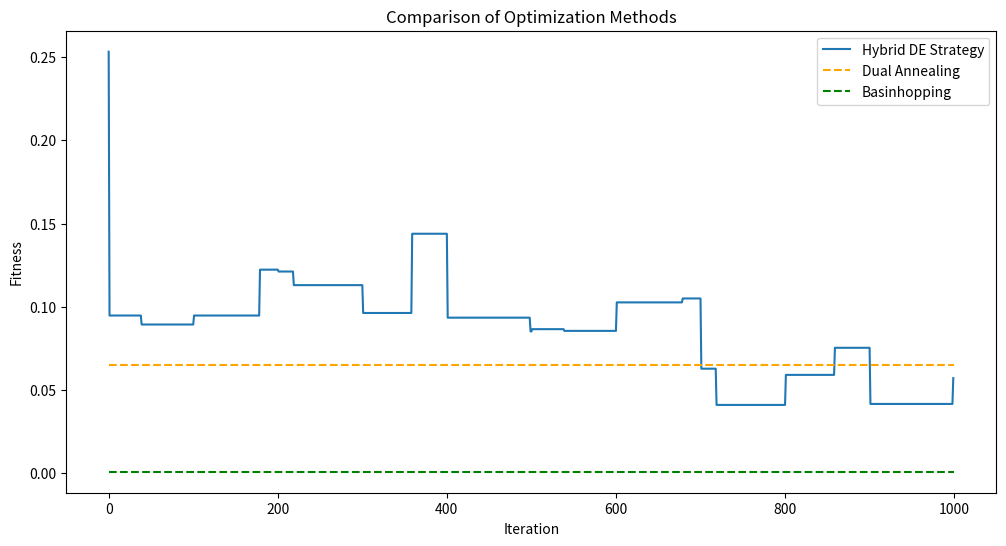}
    \caption{Comparison of Hybrid DE Strategy, Dual Annealing, and Basinhopping Techniques over 1000 Iterations}
    \label{fig:comparisonwithotherAlgorithms}
\end{figure}

The final evaluation is the comparison of the algorithm within the framework with other similar algorithms as shown in Figure \ref{fig:comparisonwithotherAlgorithms} shows these evaluation results.

The results in Figure \ref{fig:comparisonwithotherAlgorithms} show the performance of three different optimization algorithms over a series of iterations, from 0 to 1000. The y-axis measures fitness, presumably with lower values indicating better solutions. The "Hybrid DE Strategy" represents the solid blue line and displays the most fluctuation, suggesting a more exploratory search that occasionally finds better solutions. The Hybrid DE Strategy line plot provides insight into how the algorithm's fitness evolves over time, reflecting its adaptability to the dynamic problem. In contrast, the horizontal lines for Dual Annealing and Basinhopping indicate their respective best-found solutions, offering a benchmark for comparison. The "Dual Annealing" method, indicated by the dashed orange line, quickly finds a low fitness value and maintains it steadily throughout the iterations, indicating rapid convergence and a potential lack of further exploration after finding a good solution. "Basinhopping," shown with the dashed-dot green line, demonstrates the best performance with the lowest fitness values throughout, suggesting consistent and stable convergence to an optimal solution. However, the static nature of the Dual Annealing and Basinhopping results provides a baseline for evaluating the effectiveness of more dynamic approaches such as the Hybrid DE Strategy. The static nature of these algorithms makes them unsuitable for dynamic optimization problems, which shows the effectiveness of the current approach.

\section{Conclusion}\label{Conclusion}
This paper presents the Adaptive Metaheuristic Framework (AMF), a novel approach to tackling the complexities of dynamic optimization problems. The AMF, which combines differential evolution (DE) and a sophisticated adaptation module, significantly advances computational optimization. Its dynamic problem representation and real-time adaptation abilities allow it to optimize where problem parameters, constraints, and objectives constantly change. The AMF framework is unique for detecting changes in an ever-changing environment and adapting its search techniques accordingly. This adaptability increases the effectiveness of the framework and ensures that the solutions provided are relevant and useful in continuously evolving scenarios. Empirical tests conducted on dynamic optimization problems have shown that the AMF performs remarkably well, especially in preserving high-quality solutions in the face of frequent and unpredictable changes.

The AMF framework is an example of the potential of adaptive algorithms in metaheuristic optimization that combines with the adaptation components. This framework is a significant step forward in following algorithms focused on finding solutions and is also adaptive, robust, and capable of handling the complexities of real-world optimization problems. Further research can be done in this direction to refine the adaptation mechanisms, investigate the use of AMF in different industrial and real-world contexts, and broaden its capabilities to solve multi-objective and constrained optimization problems. Simply, the dynamic optimization problem generator can be replaced with any dynamic optimization use case.

\bibliographystyle{unsrturl}
\bibliography{refs.bib}

\begin{thebibliography}{10}

\bibitem{amaran2016simulation}
Satyajith Amaran, Nikolaos~V Sahinidis, Bikram Sharda, and Scott~J Bury.
\newblock Simulation optimization: a review of algorithms and applications.
\newblock {\em Annals of Operations Research}, 240:351--380, 2016.

\bibitem{Boks2021}
Rick Boks.
\newblock Dynamic configuration of operators and parameters in differential
  evolution through combined fitness and diversity-driven adaptation methods.
\newblock Master's thesis, Leiden Institute of Advanced Computer Science
  (LIACS), Leiden University, 2021.

\bibitem{math10224297}
Aleksei Vakhnin, Evgenii Sopov, and Eugene Semenkin.
\newblock On improving adaptive problem decomposition using differential
  evolution for large-scale optimization problems.
\newblock {\em Mathematics}, 10(22), 2022.
\newblock URL: \url{https://www.mdpi.com/2227-7390/10/22/4297}, \href
  {http://dx.doi.org/10.3390/math10224297} {\path{doi:10.3390/math10224297}}.

\bibitem{Martins2021}
André~Ferreira Martins, Marcelo Machado, Helena~Sofia Bernardino, et~al.
\newblock A comparative analysis of metaheuristics applied to adaptive
  curriculum sequencing.
\newblock {\em Soft Computing}, 25:11019--11034, 2021.
\newblock URL: \url{https://doi.org/10.1007/s00500-021-05836-9}, \href
  {http://dx.doi.org/10.1007/s00500-021-05836-9}
  {\path{doi:10.1007/s00500-021-05836-9}}.

\bibitem{Penas2017}
David~R. Penas, David Henriques, Patricia González, Ramón Doallo, Julio
  Saez-Rodriguez, and Julio~R. Banga.
\newblock A parallel metaheuristic for large mixed-integer dynamic optimization
  problems, with applications in computational biology.
\newblock {\em PLOS ONE}, 12(8):1--32, 08 2017.
\newblock URL: \url{https://doi.org/10.1371/journal.pone.0182186}, \href
  {http://dx.doi.org/10.1371/journal.pone.0182186}
  {\path{doi:10.1371/journal.pone.0182186}}.

\bibitem{Valdez2021}
F.~Valdez, O.~Castillo, and P.~Melin.
\newblock Bio-inspired algorithms and its applications for optimization in
  fuzzy clustering.
\newblock {\em Algorithms}, 14(4):122, 2021.
\newblock \href {http://dx.doi.org/10.3390/a14040122}
  {\path{doi:10.3390/a14040122}}.

\bibitem{Oladipo2020}
Stephen Oladipo, Yanxia Sun, and Zenghui Wang.
\newblock Optimization of pid controller with metaheuristic algorithms for dc
  motor drives: Review.
\newblock {\em International Review of Electrical Engineering (IREE)},
  15(5):18688, 2020.
\newblock \href {http://dx.doi.org/10.15866/iree.v15i5.18688}
  {\path{doi:10.15866/iree.v15i5.18688}}.

\bibitem{Zitouni2022}
Farouq Zitouni, S.~Harous, Abdelghani Belkeram, et~al.
\newblock The archerfish hunting optimizer: A novel metaheuristic algorithm for
  global optimization.
\newblock {\em Arabian Journal of Science and Engineering}, 47:2513--2553,
  2022.
\newblock \href {http://dx.doi.org/10.1007/s13369-021-06208-z}
  {\path{doi:10.1007/s13369-021-06208-z}}.

\bibitem{sahib2016new}
Mouayad~A Sahib and Bestoun~S Ahmed.
\newblock A new multiobjective performance criterion used in pid tuning
  optimization algorithms.
\newblock {\em Journal of advanced research}, 7(1):125--134, 2016.

\bibitem{Osaba2021}
E.~Osaba, Esther Villar-Rodriguez, J.~Ser, Antonio~J. Nebro, D.~Molina,
  A.~Latorre, P.~Suganthan, C.~Coello, and Francisco Herrera.
\newblock A tutorial on the design, experimentation and application of
  metaheuristic algorithms to real-world optimization problems.
\newblock {\em Swarm and Evolutionary Computation}, 64:100888, 2021.
\newblock \href {http://dx.doi.org/10.1016/j.swevo.2021.100888}
  {\path{doi:10.1016/j.swevo.2021.100888}}.

\bibitem{Uzor2014}
C.~J. Uzor, M.~Gongora, S.~Coupland, and B.~N. Passow.
\newblock Real-world dynamic optimization using an adaptive-mutation compact
  genetic algorithm.
\newblock In {\em 2014 IEEE Symposium on Computational Intelligence in Dynamic
  and Uncertain Environments (CIDUE)}, pages 1--8. IEEE, 2014.
\newblock \href {http://dx.doi.org/10.1109/CIDUE.2014.7007862}
  {\path{doi:10.1109/CIDUE.2014.7007862}}.

\bibitem{Rajwar2023}
K.~Rajwar, K.~Deep, and S.~Das.
\newblock An exhaustive review of the metaheuristic algorithms for search and
  optimization: taxonomy, applications, and open challenges.
\newblock {\em Artificial Intelligence Review}, 2023.
\newblock \href {http://dx.doi.org/10.1007/s10462-023-10470-y}
  {\path{doi:10.1007/s10462-023-10470-y}}.

\bibitem{Game2020}
P.~Game, V.~Vaze, and M.~Emmanuel.
\newblock Bio-inspired optimization: metaheuristic algorithms for optimization,
  2020.
\newblock \href {http://arxiv.org/abs/2003.11637} {\path{arXiv:2003.11637}}.

\bibitem{Kumar2020}
A.~Kumar, S.~Das, and I.~Zelinka.
\newblock A self-adaptive spherical search algorithm for real-world constrained
  optimization problems.
\newblock In {\em Proceedings of the 2020 Genetic and Evolutionary Computation
  Conference Companion}, pages 53--54. Association for Computing Machinery,
  2020.
\newblock \href {http://dx.doi.org/10.1145/3377929.3398186}
  {\path{doi:10.1145/3377929.3398186}}.

\bibitem{Yi2017}
H.-J. Yi.
\newblock Hybrid metaheuristic experiments of real-time adaptive optimization
  of parametric shading design through remote data transfer.
\newblock In {\em 2017 Winter Simulation Conference (WSC)}, pages 1--12. IEEE,
  2017.
\newblock \href {http://dx.doi.org/10.1109/WSC.2017.8247978}
  {\path{doi:10.1109/WSC.2017.8247978}}.

\bibitem{storn1997differential}
Rainer Storn and Kenneth Price.
\newblock Differential evolution--a simple and efficient heuristic for global
  optimization over continuous spaces.
\newblock {\em Journal of global optimization}, 11:341--359, 1997.

\bibitem{narloch2017diversification}
Pedro~Henrique Narloch and Rafael~Stubs Parpinelli.
\newblock Diversification strategies in differential evolution algorithm to
  solve the protein structure prediction problem.
\newblock In {\em Intelligent Systems Design and Applications: 16th
  International Conference on Intelligent Systems Design and Applications (ISDA
  2016) held in Porto, Portugal, December 16-18, 2016}, pages 125--134.
  Springer, 2017.

\end{thebibliography}

\end{document}